\documentclass{article}

\usepackage[final, nonatbib]{neurips_2023}

\usepackage{amsmath,amsfonts,bm}

\def\eqref#1{equation~\ref{#1}}

\def\1{\bm{1}}

\def\mI{{\bm{I}}}

\def\mU{{\bm{U}}}

\DeclareMathAlphabet{\mathsfit}{\encodingdefault}{\sfdefault}{m}{sl}
\SetMathAlphabet{\mathsfit}{bold}{\encodingdefault}{\sfdefault}{bx}{n}

\def\gR{{\mathcal{R}}}

\usepackage[utf8]{inputenc} 
\usepackage[T1]{fontenc}    
\usepackage{hyperref}       
\usepackage{url}            
\usepackage{booktabs}       
\usepackage{amsfonts}       
\usepackage{nicefrac}       
\usepackage{microtype}      
\usepackage{xcolor}         
\usepackage{graphicx}

\title{Estimating the Generalization in Deep Neural Networks via Sparsity}

\author{%
Yang Zhao\qquad \qquad \quad  Hao Zhang\qquad  \qquad \quad Xiuyuan Hu \\
  Department of Electronic Engineering\\
  Tsinghua University\\
  \texttt{\{zhao-yang, haozhang\}@tsinghua.edu.cn, huxy22@mails.tsinghua.edu.cn} \\
}

\begin{document}

\maketitle

\begin{abstract}
    Generalization is the key capability for deep neural networks (DNNs). However, it is challenging to give a reliable measure of the generalization ability of a DNN via only its nature. In this paper, we propose a novel method for estimating the generalization gap based on network sparsity. Two key sparsity quantities are extracted from the training results alone, which could present close relationship with model generalization. Then a simple linear model involving two key quantities are constructed to give accurate estimation of the generalization gap. By training DNNs with a wide range of generalization gap on popular datasets, we show that our key quantities and linear model could be efficient tools for estimating the generalization gap of DNNs.
\end{abstract}

\section{Introduction}

Deep neural networks (DNNs) have achieved great success in many real-world tasks \cite{DBLP:conf/cvpr/HeZRS16,DBLP:journals/corr/SimonyanZ14a, DBLP:conf/iclr/DosovitskiyB0WZ21, DBLP:journals/corr/RenHG015, DBLP:conf/cvpr/RedmonDGF16}, owing to their extraordinary generalization ability on unseen data by training with finite samples, even though they could be heavily overparameterized compared to the number of training samples. But meanwhile, due to their excessive capacity, DNNs are demonstrated to easily fit training data with arbitrary random labels or even pure noise, obviously without generalizing \cite{DBLP:conf/iclr/ZhangBHRV17}. So, a critical question raises: given a DNN, how could we estimate its generalization ability via only its nature?

Many works strive to unveil which attributes would have an underlying impact on the generalization or emerge in a generalized DNN. This may generally include the complexity or capacity, the stability or the robustness, and the sparsity of DNNs. Particularly, some of these works try to evaluate the generalization bounds that is as tight as possible while also hoping to demystify the mechanism of generalization. Despite somehow providing potential solutions on the mentioned problem, currently there remains barriers between the generalization bounds and how they could give precise estimation of the generalization in practice. Further, margin-based works leverage the prior knowledge to solve this problem, but due to the unbounded scales, they fail to give acceptable results at times. 

In this paper, we quantitatively investigate the network generalization from the perspective of a particular nature of DNNs, i.e., the sparsity of network units. Two key quantities with close relation with both DNN sparsity and generalization are extracted from the training results. They have strict bounds within an appropriate range to ensure the accurate estimation of network generalization. A practical linear model for estimating the generalization gap of DNNs is built by the two proposed quantities. We empirically found that units in DNNs trained on real data exhibit decreasing sparsity as the fraction of corrupted labels increases. By investigating several DNN models with a wide range of generalization gap, we found that both of the two proposed quantities are highly correlated with generalization gap in an approximately linear manner. This ensures satisfactory results when performing estimation on practical networks using training samples and appropriate linear model. With extensive experiments on various datasets, we show that our linear model could give a reliable estimation of the generalization gap in DNNs and better results compared to the margin-based method.

\section{Related Works}

For generalization estimation in DNNs, a conventional line is that generalization should be bounded based on certain measurements of the model complexity or capacity of DNNs where VC dimension \cite{bartlett1998sample, koltchinskii2002empirical} and Rademacher complexity \cite{bartlett2002rademacher,neyshabur2015norm, sun2016depth} are typically used. However, this approach appears unreasonable as \cite{DBLP:conf/iclr/ZhangBHRV17} show that DNNs are able to fit any possible labels, regardless of whether or not regularization techniques are employed.

Afterward, bounding the generalization based on stability or robustness seem to have received more attention. As for stability, it investigates the change of outputs when perturbing inputs or models. Generally, for keeping the network stable, generalized DNNs are expected to stay in the flat landscape neighboring the minima \cite{DBLP:journals/neco/HochreiterS97a, DBLP:conf/iclr/KeskarMNST17, DBLP:conf/colt/GonenS17, foret2020sharpness, zhao2022penalizing}. However, it is still controversial in regards to how to appropriately gauge the "flatness" of a minimum. \cite{DBLP:conf/nips/NeyshaburBMS17} argue that the definition provided by \cite{DBLP:conf/iclr/KeskarMNST17} could not well capture the generalization behavior. In addition, \cite{DBLP:conf/icml/DinhPBB17} point that sharp minima may also leads to the generalization when using different definitions of flatness. In contrast to stability, robustness investigates the variation of outputs with respect to the input space. One typical evaluation of robustness is the margin of the predictions, where generalized DNNs are supposed to have large margin to ensure the prediction robustness \cite{DBLP:journals/tsp/SokolicGSR17, DBLP:conf/nips/ElsayedKMRB18}. In particular, \cite{DBLP:conf/nips/BartlettFT17} use the margin distribution of the outputs and normalized it with the spectral norm. Based on this margin distribution, \cite{DBLP:conf/iclr/JiangKMB19} further use the margin distribution of hidden layers to give an estimation of generalization. But meanwhile, \cite{DBLP:conf/icml/Arora0NZ18} argue that methods in \cite{DBLP:conf/nips/BartlettFT17,DBLP:conf/iclr/NeyshaburBS18} could not yet give  bounds of sample complexity better than naive parameter counting.
 
Sparsity of the network unit is considered as an important sign that units present highly specialized \cite{DBLP:journals/corr/abs-1806-02891} and can perceive the intrinsic natural features to provide basis for generalization on unseen data. This is particularly significant for CNN units since they are found to be conceptually aligned with our vision cognition \cite{DBLP:journals/pnas/BauZSLZ020,DBLP:journals/corr/FongV17}. Trained units can present specific natural concepts that gives disentangled representations for different classes \cite{DBLP:conf/cvpr/BauZKO017, DBLP:journals/pnas/BauZSLZ020}. \cite{DBLP:journals/corr/abs-1806-02891} show that for CNNs, only several conceptual related units are necessary for each category, and \cite{DBLP:conf/iclr/MorcosBRB18, DBLP:conf/iclr/LeavittM21} claim that generalization should not rely on the single direction of the units. Based on empirical observations, sparsity over networks is found to be very helpful for generalization \cite{DBLP:conf/nips/BartoldsonMBE20, DBLP:conf/ijcai/Liu20a, DBLP:journals/corr/abs-1906-11626}. So methods such as dropout regularization are proposed for inducing the DNNs to become sparse during training \cite{srivastava2014dropout, DBLP:journals/ijon/ScardapaneCHU17, DBLP:conf/nips/WenWWCL16, DBLP:journals/corr/abs-1712-01312, DBLP:journals/corr/AchilleS17}.


\section{Method}

DNNs are bio-inspired artificial neural networks that generally consist of multiple layers where each layer is a collection of certain amount of units. Similar to neuroscience, a unit in DNNs\footnote{For CNNs, it refers to the activated feature map outputted by the corresponding convolutional filter} generally behaves as a perceptive node, and makes responses to the inputs. Given an input $\mI$, the unit $\mU$ is expected to present specific function preference and would be highly responded if it could capture the feature of this input, $f : \mI \rightarrow \mU$.

Consider a classification problem with a \textbf{training} dataset $\mathcal{D}$ including $N$ classes $\{\mathcal{D}_k\}^{N}_{k=1}$, $\mathcal{D} = \mathcal{D}_1 \cup \cdots \cup \mathcal{D}_{N}$. To obtain excellent generalization on unseen data, units are trained to be functionally specialized to perceive diverse intrinsic natural features hidden in a dataset. For a single class, only a small group of units, which are function-correlated to the common features of this class, would be highly active. Therefore, this results in the \emph{sparsity} of units. On the contrary, due to the its excessive capacity, it is possible for DNN to just "memorize" each sample in the class and then simply "mark" it as belonging to this class. In this situation, much more units are needed to be active since they could not give true perception in regards to class-specific intrinsic features. In this way, we would begin with identifying the group of units which are highly active on samples in the specific class $\mathcal{D}_j$.

\subsection{Cumulative Unit Ablation}

In general, unit ablation operation is supposed to be able to make connections between DNN sparsity and generalization. Single unit ablation individually checks each unit that would cause a performance deterioration when being removed\footnote{Removing or ablating a unit is generally implemented via forcing the elements in this unit to be all deactivated. For example, for ReLU activation, a unit is assigned to all zeros if being removed} from the DNN. However, this tells nearly no information on the collaborative effect of a group of units. Here, we introduce the cumulative unit ablation for studying the group effect of units to the given network model. In cumulative unit ablation, units at a given layer are first arranged into a list $\gR$ which is ranked by the quantity of certain attribute of a unit,
\begin{equation}
    \gR = < h(\mU_0), h(\mU_1), \cdots, h(\mU_i) >
\end{equation}
where $< \cdot >$ is the sorted list and $h(\mU_i)$ denotes a given attribute of the unit $\mU_i$, such as L1-norm \cite{DBLP:journals/corr/abs-1806-02891}, class selectivity \cite{DBLP:conf/iclr/MorcosBRB18}, topological entropy \cite{zhao2021quantitative} etc. Since we are focusing on the group of units that are highly active to $\mathcal{D}_j$, we use the L1-norm value of the unit as the target attribute $h(\mU_i)$ in the cumulative ablation, i.e. $h(\mU_i) = \sum_{x, y} \mU_i(x, y)$. In this way, units in the list $\gR$ would be ordered based on its L1-norm value $h(\mU_i)$. 

Then after ranking, units are removed progressively from the head unit to the end unit in this ordered list $\gR$, and in the meantime, the evaluation of performance is recorded as a characterization of the growing effect of this attribute to the DNN.

During our implementation, we would perform the cumulative unit ablation on $\mathcal{D}_j$ twice, separately according to two different ordered lists. One is $\gR$, sorted by the descending rank of $h(\mU_i)$, and the other one is $\gR_r$, sorted by the ascending rank of $h(\mU_i)$. Correspondingly, the two evolution curves of network accuracy with respect to the number of units being removed (notated as $n$) could be recorded, where $E(n, \mathcal{D}_j)$ denotes the accuracy evolution on the ascending rank and $E_r(n, \mathcal{D}_j)$ denotes the other. Fig.1A illustrates typically $E(n, \mathcal{D}_j)$ and $E_r(n, \mathcal{D}_j)$.

Notably, compared to units at the shallow layers, units in the deeper layers are considered to perceive "high-level" class-related features \cite{DBLP:journals/pnas/BauZSLZ020} and give more representative of the specific class. Units in deeper layers tend to be more sparse, which means that cumulative unit ablation would present more significant effect. So practical implementations could focus on the deeper layers to give better results.



\begin{figure}[t]
    \centering
    \vspace{-0.1in}
    \includegraphics[width = 0.9\columnwidth]{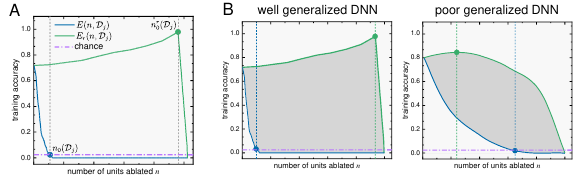}
    \caption{(A) Example of $E(n, \mathcal{D}_j)$ (blue) and $E_r(n, \mathcal{D}_j)$ (green), where the two markers represent for the two turning points $n_0(\mathcal{D}_j)$ and $n_0(\mathcal{D}_j)$. (B) The turning points and the enclosed areas (painted in gray) regarding DNNs with well generalization (left) and poor generalization (right).}
\end{figure}

\subsection{Turning Points}

With the ordered list in descending rank $\gR$ on $\mathcal{D}_j$, highly active units are removed at the beginning of the evolution process. So the accuracy experiences a continuous decrease since the neural network gradually lose its function on extracting the features in this class. Notably, the accuracy may reach below chance level $acc_{chance}$ after some critical units are removed and then remain this situation with little variation until all units are removed. We mark the minimum number of removed units causing a complete damage to the function for the DNN on the dataset $\mathcal{D}_j$ as the \emph{turning point} of $E(n, \mathcal{D}_j)$,
\begin{equation}
    n_0(\mathcal{D}_j) = \inf \{n \vert E(n, \mathcal{D}_j) \leq acc_{chance}\}
\end{equation}
\noindent Apparently, if $n_0(\mathcal{D}_j)$ is large, it means that the majority of units have positive contribution to $\mathcal{D}_j$, so it requires the deactivation of more critical units to completely lose response to $\mathcal{D}_j$.

On the contrary, with the ordered list in ascending rank $\gR_r$, units that are highly active to $\mathcal{D}_j$ would be preserved at the beginning and removed near the end of the evolution process. So the corresponding $E_r(n, \mathcal{D}_j)$ generally experiences a continual slight increase at the early stage of evolution, and this would keep in the most time during the evolution until the accuracy reaches at the maximum. After this point, the accuracy would drop abruptly to below chance level. Similarly, we mark the maximal point as the turning point of $E_r(n, \mathcal{D}_j)$,
\begin{equation}
     n_{0}^{r}(\mathcal{D}_j) = \mathop{\arg\max}_{n} \ E_r(n, \mathcal{D}_j)
\end{equation}
\noindent Notably, $M - n_{0}^{r}(\mathcal{D}_j)$ represents the minimum number of units being activated jointly that could give the most performance. If $n_{0}^{r}(\mathcal{D}_j)$ is large, it means that for $\mathcal{D}_j$, most units are unrelated in function and activating only a small number of critical units would be able to provide the best effect.


\subsection{Key Quantities and Estimation of Generalization Gap in DNNs}

From previous demonstration, units that are highly active to $\mathcal{D}_j$ should be more sparse for DNNs with better generalization. This means that performance would be more sensitive to the removals of these "important" units during ablation for well generalized DNNs. In other words, both $E(n, \mathcal{D}_j)$ and $E_r(n, \mathcal{D}_j)$ are more steep than those in "memorized" networks, as shown in Fig.1B.


For DNNs with better generalization, it is expected that $n_{0}(\mathcal{D}_j)$ should be smaller while $n_{0}^{r}(\mathcal{D}_j)$ should be larger. So we could simply combine the two values,
\begin{equation}
    \zeta(\mathcal{D}_j) = \frac{n_0(\mathcal{D}_j) + M - n_{0}^{r}(\mathcal{D}_j)}{2M}
\end{equation}
\noindent where $\zeta(\mathcal{D}_j) \in [0, 1]$. The smaller the value is, the sparser the critical units are in the DNN on the classification of $\mathcal{D}_j$. This is one of the key quantity derived from cumulative unit ablation.


On the other hand, impacted by the two turning points, the area enclosed by the two curves $E(n, \mathcal{D}_j)$ and $E_r(n, \mathcal{D}_j)$ should be also valuable. In addition to $\zeta(\mathcal{D}_j)$, we have another key quantity,
\begin{equation}
    \kappa(\mathcal{D}_j) = \frac{1}{M}\sum_{n=0}^{M} \vert E_r(n, \mathcal{D}_j) - E(n, \mathcal{D}_j) \vert
\end{equation}
\noindent Here, the coefficient $\frac{1}{M}$ is used to adjust the area value to fall in the range from 0 to 1. Opposite to $\zeta(\mathcal{D}_j)$, the larger $\kappa(\mathcal{D}_j)$ indicates the critical units are sparser in the DNN on the classification of $\mathcal{D}_j$.



For dataset $\{\mathcal{D}_k\}_{k=1}^N$ of all classes, we could simply use the average to make fusion for the two characterization on various data classes to achieve the ensemble effect,
\begin{equation}
        \label{eqn: calculate genelized rate}
    \phi(\mathcal{D}) = \frac{1}{N}\sum_{j=1}^N \phi(\mathcal{D}_j), ~~ \phi \in \left\{ \zeta, \kappa \right\}
\end{equation}


The $\zeta(\mathcal{D})$ and $\kappa(\mathcal{D})$ are in high correlation to the generalization of DNN. In fact, they could be utilized to estimate the generalization ability via a simple linear model,
\begin{equation}
    \hat{g}(\mathcal{D}) = a \cdot \zeta(\mathcal{D}) + b \cdot \kappa(\mathcal{D}) + c
    \label{eqn: estimation}
\end{equation}
where $a$, $b$ and $c$ are the parameters of this linear model. During our investigations, we find a simple linear model is sufficient in this situation since the two quantities and the generalization present a highly negative linear relationship. It should be noted that for clarity, in the following sections we will use the word "model" to refer to this linear model that predicts the generalization gap of DNNs, and use the word "network" to refer to trained neural network models.


\section{Experimental Results}

In this section, we are going to implement our method on the classification task of CIFAR100 by using VGG16 architecture \cite{DBLP:journals/corr/SimonyanZ14a}. Additional results by using other network architectures and ImageNet dataset \cite{deng2009imagenet} could be found in the Appendix section.

\paragraph{Experiment Setting}
For obtaining networks with a wide range of generalization gap, we randomly corrupt the labels with certain percentage for each class in the dataset as \cite{DBLP:conf/iclr/ZhangBHRV17}. Meanwhile, we also use different training recipes to obtain the networks with diverse generalization gap, such as varying the common regularizations like weight decay, batch normalization, dropout, etc. All the networks are trained to reach at almost 1.0 accuracy on training set. We have trained 80 networks in total and their generalization errors (accuracy on testing set) are ranged from 0.294 to 0.987.

\begin{figure*}[t]
    \centering
    \vspace{-0.15in}
    \includegraphics[width = 1\columnwidth]{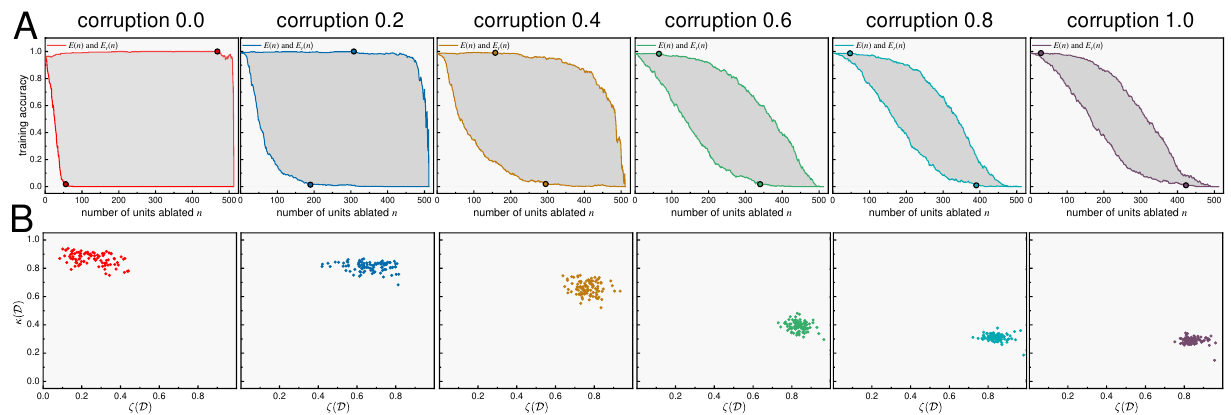}
    \vspace{-0.2in}
    \caption{Results of networks trained via datasets with partially randomized labels. (A) The evolution example curves of accuracy $E(n, \mathcal{D}_j)$ and $E_r(n, \mathcal{D}_j)$ on a single class. (B) Scatter plot between the two quantities $\zeta(\mathcal{D})$ and $\kappa(\mathcal{D})$ across all the classes in the separate corrupted datasets.}
    \label{fig: random_label}
    \vspace{-0.05in}
\end{figure*}

\paragraph{Key quantities of networks trained with partially randomized labels.}
We calculate the two key quantities of networks that are trained with exactly the same training strategy but on datasets with different percentage of randomized labels.

Firstly, we perform the cumulative unit ablation for the networks on their training dataset. Fig.\ref{fig: random_label}A shows the two evolution curves $E(n, \mathcal{D}_j)$ and $E_r(n, \mathcal{D}_j)$ on the same class from the datasets separately with 0, 0.2, 0.4, 0.6, 0.8, 1.0 fractions of randomized labels. In the figure, markers with a black border denote the two turning points and the area enclosed by the two curves is painted with gray. We could see that as the fraction of randomized labels goes higher, the first turning point $n_0(\mathcal{D}_j)$ gradually increase while the second turning point $n_0^{r}(\mathcal{D}_j)$ decrease. In addition to the two turning points, the area becomes smaller as well.

Then, we calculate the two quantities $\zeta(\mathcal{D})$ and $\kappa(\mathcal{D})$ for these networks on all the classes. Fig.\ref{fig: random_label}B makes the scatter plot of the point pair $\left(\zeta(\mathcal{D}_j), \kappa(\mathcal{D}_j)\right)$ for all the 100 classes. According to previous demonstration, for better generalization, $\zeta(\mathcal{D})$ should be smaller while $\kappa(\mathcal{D})$ should be larger. This makes that the quantity pair should locate around the top left corner of the scatter figure. As expected, the point group moves from the top left regularly to right bottom corner as the fraction of label corruption increases. This confirms that when networks are trained with partially randomized labels, the two sparsity quantities could effectively indicate the generalization ability of these networks.


\paragraph{Estimating the generalization of trained networks.} After the calculation of the two sparsity quantities $\zeta(\mathcal{D})$ and $\kappa(\mathcal{D})$ for all 80 the trained networks, we are going to estimate their generalization.

\begin{figure*}[t]
    \centering
    \vspace{-0.2in}
    \includegraphics[width = 0.9\columnwidth]{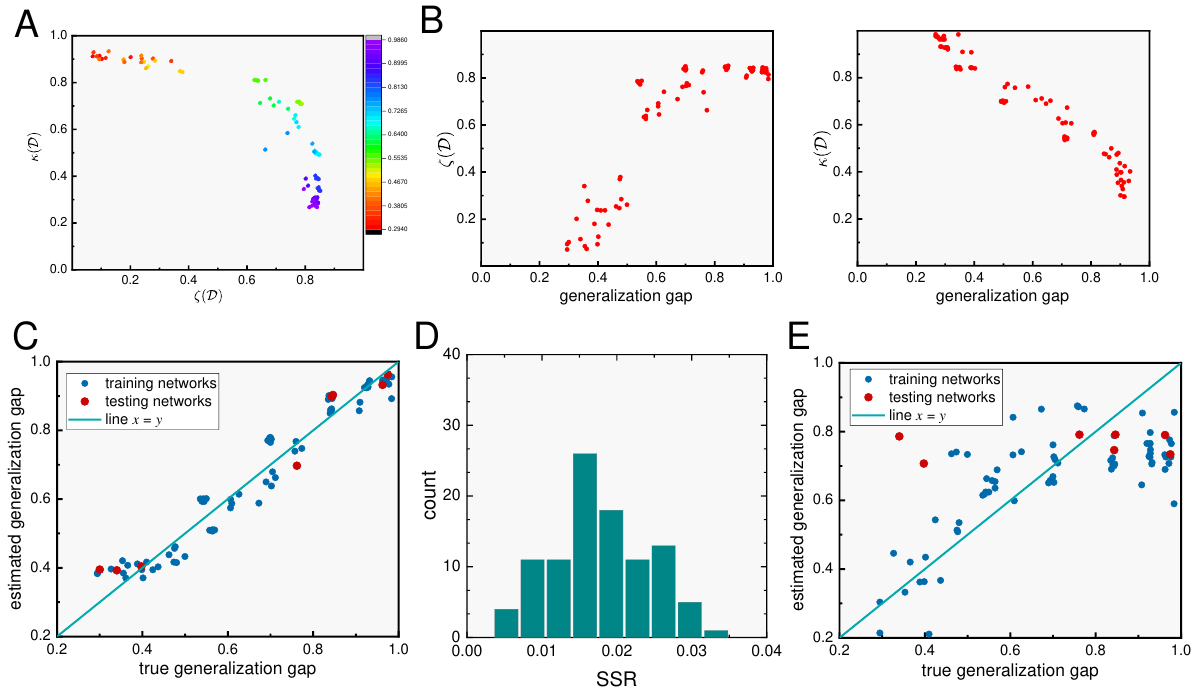}
    \vspace{-0.05in}
    \caption{(A) Scatter plot between the two quantities $\zeta(\mathcal{D})$ and $\kappa(\mathcal{D})$. (B) Scatter plots of generalization gap separately with respect to $\zeta(\mathcal{D})$ (left) and $\kappa(\mathcal{D})$ (right). (C) Scatter plot between the estimated generalization gap and the true generalization gap, where the blue points denote the training networks and the red points denote testing networks. (D) Histogram of SSRs of the 100 repeated tests. (E) Scatter plot between the true generalization gap and that estimated by margin-based method.}
    \label{fig: curve fitting}
    \vspace{-0.08in}
  \end{figure*}

We begin with marking the quantity pair $(\zeta(\mathcal{D}), \kappa(\mathcal{D}))$ of each network with a scatter plot, as shown in Fig.\ref{fig: curve fitting}A. In this figure, the colors of points vary progressively from red to purple, which indicates the true generalization gap of networks from small to large. As expected, the point of networks with better generalization ability mostly lie in the top right corner of figure while these with poor generalization ability lie in the bottom left corner. In the meantime, we could clearly find that the two quantities $\zeta(\mathcal{D})$ and $\kappa(\mathcal{D})$ are negatively correlated, but not in a linear manner. 

Then, we gives two scatters plots in the Fig.\ref{fig: curve fitting}B, one for $\zeta(\mathcal{D})$ and generalization gap (left) and the other one for $\kappa(\mathcal{D})$ and generalization gap (right). We could see that as the generalization gap goes higher, the $\zeta(\mathcal{D})$ increase while the $\kappa(\mathcal{D})$ decrease. This confirms that both of the two quantities could indeed provide efficient indication of the generalization ability of networks.

Next, we randomly split all the 80 trained networks into two sets by fractions of 0.9 and 0.1. The set with 72 networks is used as the training networks to build the linear model for estimating the generalization gap via Eq.\ref{eqn: estimation}, and the other set with 8 networks is used as a testing networks to check the effectiveness of this linear model after fitting. Clearly, the estimation here is a typical linear regression problem and could simply be solved by using least square. 

Fig.\ref{fig: curve fitting}C(1) illustrates the results of linear fitting, where $y = x$ is the reference line for the perfect fitting. We could see that the training points scattered closely to the reference line, showing that the two key quantities and the generalization gap may be highly linearly correlated. The Pearson correlation coefficient between the estimated generalization gaps and the true values of the training networks is 0.979, confirming their linear relationship. For testing, the fitted linear model performs well on these networks in the testing set. In addition, we use the summation of squared residuals (SSR) \cite{archdeacon1994correlation} as a yardstick for checking the predicting effect on testing networks. SSR is a conventional measurement of the performance of fitted models and usually used for model selection. In our case, it is $0.023$, which is very small and indicates this model is ready to give excellent prediction in practice.

For checking the stability of estimation when using our method, we repeat the previous estimation 100 times and each time use a new splitted dataset but still with the same fraction. Fig.\ref{fig: curve fitting}D presents the statistical results of RSS with respect to all the testing sets. For the 100 splits, all the RSSs keep in a low value below $0.035$. This verifies the overall validity of two quantities $\zeta(\mathcal{D})$ and $\kappa(\mathcal{D})$ to be indicators of generalization ability and the effectiveness of linear model for generalization estimation.

\paragraph{Comparison with margin-based estimation.} Lastly, we give comparisons of our method with margin-based method proposed in \cite{DBLP:conf/iclr/JiangKMB19}. Typically, this method collects the margin distribution on a given dataset and use key features of this distribution as arguments for fitting the generalization gap.

Here, we keep previous setups and the dataset being the same as used in Fig.\ref{fig: curve fitting}C. Fig.\ref{fig: curve fitting}E shows the corresponding estimation results. We could see that although the margin based model could estimate the generalization gap to some extent, it presents with worse linear correlation (Pearson correlation coefficient is $0.75$) than our method. When predicting the generalization gap of testing networks, it presents a larger SSR, which is almost $0.5$.


We suppose that two possible factors might lead to the errors via margin-based method. One is that due to the non-linearity, margins in DNNs are actually intractable. Currently, the distance of a sample to the margin is approximately acquired by using the first-order Taylor approximation \cite{DBLP:conf/nips/ElsayedKMRB18}, which will bring error into the estimation. The second factor is that the calculated margins are not bounded in scales. This may lead to that for different models, their margins can differ by orders of magnitude considering the distinct training settings. In this way, the corresponding linear model may be ill-conditioned. During our implementation, we found that for some cases (especially for the networks trained with batch normalization), this problem becomes even worse.

\section{Conclusion}

We propose a method for reliably estimating the generalization ability of DNNs, where we use specific-designed cumulative unit ablation to capture two key sparsity quantities. These quantities are found strong linearly correlated with generalization ability, so a linear model is built for generalization estimation. Extensive experiments show that our linear model can give accurate estimation in practice.

\small{
\bibliographystyle{plain}
\bibliography{egbib}

\begin{thebibliography}{10}

\bibitem{DBLP:journals/corr/AchilleS17}
Alessandro Achille and Stefano Soatto.
\newblock On the emergence of invariance and disentangling in deep
  representations.
\newblock {\em arXiv}, abs/1706.01350, 2017.

\bibitem{archdeacon1994correlation}
Thomas~J Archdeacon.
\newblock {\em Correlation and regression analysis: a historian's guide}.
\newblock Univ of Wisconsin Press, 1994.

\bibitem{DBLP:conf/icml/Arora0NZ18}
Sanjeev Arora, Rong Ge, Behnam Neyshabur, and Yi~Zhang.
\newblock Stronger generalization bounds for deep nets via a compression
  approach.
\newblock In Jennifer~G. Dy and Andreas Krause, editors, {\em Proceedings of
  the 35th International Conference on Machine Learning, {ICML} 2018,
  Stockholmsm{\"{a}}ssan, Stockholm, Sweden, July 10-15, 2018}, volume~80 of
  {\em Proceedings of Machine Learning Research}, pages 254--263. {PMLR}, 2018.

\bibitem{bartlett1998sample}
Peter~L Bartlett.
\newblock The sample complexity of pattern classification with neural networks:
  the size of the weights is more important than the size of the network.
\newblock {\em IEEE transactions on Information Theory}, 44(2):525--536, 1998.

\bibitem{DBLP:conf/nips/BartlettFT17}
Peter~L. Bartlett, Dylan~J. Foster, and Matus Telgarsky.
\newblock Spectrally-normalized margin bounds for neural networks.
\newblock In Isabelle Guyon, Ulrike von Luxburg, Samy Bengio, Hanna~M. Wallach,
  Rob Fergus, S.~V.~N. Vishwanathan, and Roman Garnett, editors, {\em Advances
  in Neural Information Processing Systems 30: Annual Conference on Neural
  Information Processing Systems 2017, December 4-9, 2017, Long Beach, CA,
  {USA}}, pages 6240--6249, 2017.

\bibitem{bartlett2002rademacher}
Peter~L Bartlett and Shahar Mendelson.
\newblock Rademacher and gaussian complexities: Risk bounds and structural
  results.
\newblock {\em Journal of Machine Learning Research}, 3(Nov):463--482, 2002.

\bibitem{DBLP:conf/nips/BartoldsonMBE20}
Brian Bartoldson, Ari~S. Morcos, Adrian Barbu, and Gordon Erlebacher.
\newblock The generalization-stability tradeoff in neural network pruning.
\newblock In Hugo Larochelle, Marc'Aurelio Ranzato, Raia Hadsell,
  Maria{-}Florina Balcan, and Hsuan{-}Tien Lin, editors, {\em Advances in
  Neural Information Processing Systems 33: Annual Conference on Neural
  Information Processing Systems 2020, NeurIPS 2020, December 6-12, 2020,
  virtual}, 2020.

\bibitem{DBLP:conf/cvpr/BauZKO017}
David Bau, Bolei Zhou, Aditya Khosla, Aude Oliva, and Antonio Torralba.
\newblock Network dissection: Quantifying interpretability of deep visual
  representations.
\newblock In {\em 2017 {IEEE} Conference on Computer Vision and Pattern
  Recognition, {CVPR} 2017, Honolulu, HI, USA, July 21-26, 2017}, pages
  3319--3327. {IEEE} Computer Society, 2017.

\bibitem{DBLP:journals/pnas/BauZSLZ020}
David Bau, Jun{-}Yan Zhu, Hendrik Strobelt, {\`{A}}gata Lapedriza, Bolei Zhou,
  and Antonio Torralba.
\newblock Understanding the role of individual units in a deep neural network.
\newblock {\em Proc. Natl. Acad. Sci. {USA}}, 117(48):30071--30078, 2020.

\bibitem{deng2009imagenet}
Jia Deng, Wei Dong, Richard Socher, Li-Jia Li, Kai Li, and Li~Fei-Fei.
\newblock Imagenet: A large-scale hierarchical image database.
\newblock In {\em 2009 IEEE conference on computer vision and pattern
  recognition}, pages 248--255. Ieee, 2009.

\bibitem{DBLP:conf/icml/DinhPBB17}
Laurent Dinh, Razvan Pascanu, Samy Bengio, and Yoshua Bengio.
\newblock Sharp minima can generalize for deep nets.
\newblock In Doina Precup and Yee~Whye Teh, editors, {\em Proceedings of the
  34th International Conference on Machine Learning, {ICML} 2017, Sydney, NSW,
  Australia, 6-11 August 2017}, volume~70 of {\em Proceedings of Machine
  Learning Research}, pages 1019--1028. {PMLR}, 2017.

\bibitem{DBLP:conf/iclr/DosovitskiyB0WZ21}
Alexey Dosovitskiy, Lucas Beyer, Alexander Kolesnikov, Dirk Weissenborn,
  Xiaohua Zhai, Thomas Unterthiner, Mostafa Dehghani, Matthias Minderer, Georg
  Heigold, Sylvain Gelly, Jakob Uszkoreit, and Neil Houlsby.
\newblock An image is worth 16x16 words: Transformers for image recognition at
  scale.
\newblock In {\em 9th International Conference on Learning Representations,
  {ICLR} 2021, Virtual Event, Austria, May 3-7, 2021}. OpenReview.net, 2021.

\bibitem{DBLP:conf/nips/ElsayedKMRB18}
Gamaleldin~F. Elsayed, Dilip Krishnan, Hossein Mobahi, Kevin Regan, and Samy
  Bengio.
\newblock Large margin deep networks for classification.
\newblock In Samy Bengio, Hanna~M. Wallach, Hugo Larochelle, Kristen Grauman,
  Nicol{\`{o}} Cesa{-}Bianchi, and Roman Garnett, editors, {\em Advances in
  Neural Information Processing Systems 31: Annual Conference on Neural
  Information Processing Systems 2018, NeurIPS 2018, December 3-8, 2018,
  Montr{\'{e}}al, Canada}, pages 850--860, 2018.

\bibitem{DBLP:journals/corr/FongV17}
Ruth Fong and Andrea Vedaldi.
\newblock Interpretable explanations of black boxes by meaningful perturbation.
\newblock {\em arXiv}, abs/1704.03296, 2017.

\bibitem{foret2020sharpness}
Pierre Foret, Ariel Kleiner, Hossein Mobahi, and Behnam Neyshabur.
\newblock Sharpness-aware minimization for efficiently improving
  generalization.
\newblock {\em arXiv preprint arXiv:2010.01412}, 2020.

\bibitem{DBLP:conf/colt/GonenS17}
Alon Gonen and Shai Shalev{-}Shwartz.
\newblock Fast rates for empirical risk minimization of strict saddle problems.
\newblock In Satyen Kale and Ohad Shamir, editors, {\em Proceedings of the 30th
  Conference on Learning Theory, {COLT} 2017, Amsterdam, The Netherlands, 7-10
  July 2017}, volume~65 of {\em Proceedings of Machine Learning Research},
  pages 1043--1063. {PMLR}, 2017.

\bibitem{DBLP:conf/cvpr/HeZRS16}
Kaiming He, Xiangyu Zhang, Shaoqing Ren, and Jian Sun.
\newblock Deep residual learning for image recognition.
\newblock In {\em 2016 {IEEE} Conference on Computer Vision and Pattern
  Recognition, {CVPR} 2016, Las Vegas, NV, USA, June 27-30, 2016}, pages
  770--778, 2016.

\bibitem{DBLP:journals/neco/HochreiterS97a}
Sepp Hochreiter and J{\"{u}}rgen Schmidhuber.
\newblock Flat minima.
\newblock {\em Neural Comput.}, 9(1):1--42, 1997.

\bibitem{DBLP:conf/iclr/JiangKMB19}
Yiding Jiang, Dilip Krishnan, Hossein Mobahi, and Samy Bengio.
\newblock Predicting the generalization gap in deep networks with margin
  distributions.
\newblock In {\em 7th International Conference on Learning Representations,
  {ICLR} 2019, New Orleans, LA, USA, May 6-9, 2019}, 2019.

\bibitem{DBLP:conf/iclr/KeskarMNST17}
Nitish~Shirish Keskar, Dheevatsa Mudigere, Jorge Nocedal, Mikhail Smelyanskiy,
  and Ping Tak~Peter Tang.
\newblock On large-batch training for deep learning: Generalization gap and
  sharp minima.
\newblock In {\em 5th International Conference on Learning Representations,
  {ICLR} 2017, Toulon, France, April 24-26, 2017, Conference Track
  Proceedings}, 2017.

\bibitem{koltchinskii2002empirical}
Vladimir Koltchinskii and Dmitry Panchenko.
\newblock Empirical margin distributions and bounding the generalization error
  of combined classifiers.
\newblock {\em The Annals of Statistics}, 30(1):1--50, 2002.

\bibitem{DBLP:conf/iclr/LeavittM21}
Matthew~L. Leavitt and Ari~S. Morcos.
\newblock Selectivity considered harmful: evaluating the causal impact of class
  selectivity in dnns.
\newblock In {\em 9th International Conference on Learning Representations,
  {ICLR} 2021, Virtual Event, Austria, May 3-7, 2021}. OpenReview.net, 2021.

\bibitem{DBLP:conf/ijcai/Liu20a}
Shiwei Liu.
\newblock Learning sparse neural networks for better generalization.
\newblock In Christian Bessiere, editor, {\em Proceedings of the Twenty-Ninth
  International Joint Conference on Artificial Intelligence, {IJCAI} 2020},
  pages 5190--5191. ijcai.org, 2020.

\bibitem{DBLP:journals/corr/abs-1906-11626}
Shiwei Liu, Decebal~Constantin Mocanu, and Mykola Pechenizkiy.
\newblock On improving deep learning generalization with adaptive sparse
  connectivity.
\newblock {\em arXiv preprint}, abs/1906.11626, 2019.

\bibitem{DBLP:journals/corr/abs-1712-01312}
Christos Louizos, Max Welling, and Diederik~P. Kingma.
\newblock Learning sparse neural networks through l\({}_{\mbox{0}}\)
  regularization.
\newblock {\em arXiv preprint}, abs/1712.01312, 2017.

\bibitem{DBLP:conf/iclr/MorcosBRB18}
Ari~S. Morcos, David G.~T. Barrett, Neil~C. Rabinowitz, and Matthew Botvinick.
\newblock On the importance of single directions for generalization.
\newblock In {\em 6th International Conference on Learning Representations,
  {ICLR} 2018, Vancouver, BC, Canada, April 30 - May 3, 2018, Conference Track
  Proceedings}. OpenReview.net, 2018.

\bibitem{DBLP:conf/nips/NeyshaburBMS17}
Behnam Neyshabur, Srinadh Bhojanapalli, David McAllester, and Nati Srebro.
\newblock Exploring generalization in deep learning.
\newblock In Isabelle Guyon, Ulrike von Luxburg, Samy Bengio, Hanna~M. Wallach,
  Rob Fergus, S.~V.~N. Vishwanathan, and Roman Garnett, editors, {\em Advances
  in Neural Information Processing Systems 30: Annual Conference on Neural
  Information Processing Systems 2017, December 4-9, 2017, Long Beach, CA,
  {USA}}, pages 5947--5956, 2017.

\bibitem{DBLP:conf/iclr/NeyshaburBS18}
Behnam Neyshabur, Srinadh Bhojanapalli, and Nathan Srebro.
\newblock A pac-bayesian approach to spectrally-normalized margin bounds for
  neural networks.
\newblock In {\em 6th International Conference on Learning Representations,
  {ICLR} 2018, Vancouver, BC, Canada, April 30 - May 3, 2018, Conference Track
  Proceedings}, 2018.

\bibitem{neyshabur2015norm}
Behnam Neyshabur, Ryota Tomioka, and Nathan Srebro.
\newblock Norm-based capacity control in neural networks.
\newblock In {\em Conference on Learning Theory}, pages 1376--1401. PMLR, 2015.

\bibitem{DBLP:conf/cvpr/RedmonDGF16}
Joseph Redmon, Santosh~Kumar Divvala, Ross~B. Girshick, and Ali Farhadi.
\newblock You only look once: Unified, real-time object detection.
\newblock In {\em 2016 {IEEE} Conference on Computer Vision and Pattern
  Recognition, {CVPR} 2016, Las Vegas, NV, USA, June 27-30, 2016}, pages
  779--788. {IEEE} Computer Society, 2016.

\bibitem{DBLP:journals/corr/RenHG015}
Shaoqing Ren, Kaiming He, Ross~B. Girshick, and Jian Sun.
\newblock Faster {R-CNN:} towards real-time object detection with region
  proposal networks.
\newblock {\em CoRR}, abs/1506.01497, 2015.

\bibitem{DBLP:journals/ijon/ScardapaneCHU17}
Simone Scardapane, Danilo Comminiello, Amir Hussain, and Aurelio Uncini.
\newblock Group sparse regularization for deep neural networks.
\newblock {\em Neurocomputing}, 241:81--89, 2017.

\bibitem{DBLP:journals/corr/SimonyanZ14a}
Karen Simonyan and Andrew Zisserman.
\newblock Very deep convolutional networks for large-scale image recognition.
\newblock In Yoshua Bengio and Yann LeCun, editors, {\em 3rd International
  Conference on Learning Representations, {ICLR} 2015, San Diego, CA, USA, May
  7-9, 2015, Conference Track Proceedings}, 2015.

\bibitem{DBLP:journals/tsp/SokolicGSR17}
Jure Sokolic, Raja Giryes, Guillermo Sapiro, and Miguel R.~D. Rodrigues.
\newblock Robust large margin deep neural networks.
\newblock {\em {IEEE} Trans. Signal Process.}, 65(16):4265--4280, 2017.

\bibitem{srivastava2014dropout}
Nitish Srivastava, Geoffrey Hinton, Alex Krizhevsky, Ilya Sutskever, and Ruslan
  Salakhutdinov.
\newblock Dropout: a simple way to prevent neural networks from overfitting.
\newblock {\em The journal of machine learning research}, 15(1):1929--1958,
  2014.

\bibitem{sun2016depth}
Shizhao Sun, Wei Chen, Liwei Wang, Xiaoguang Liu, and Tie-Yan Liu.
\newblock On the depth of deep neural networks: A theoretical view.
\newblock In {\em Proceedings of the AAAI Conference on Artificial
  Intelligence}, volume~30, 2016.

\bibitem{DBLP:conf/nips/WenWWCL16}
Wei Wen, Chunpeng Wu, Yandan Wang, Yiran Chen, and Hai Li.
\newblock Learning structured sparsity in deep neural networks.
\newblock In Daniel~D. Lee, Masashi Sugiyama, Ulrike von Luxburg, Isabelle
  Guyon, and Roman Garnett, editors, {\em Advances in Neural Information
  Processing Systems 29: Annual Conference on Neural Information Processing
  Systems 2016, December 5-10, 2016, Barcelona, Spain}, pages 2074--2082, 2016.

\bibitem{DBLP:conf/iclr/ZhangBHRV17}
Chiyuan Zhang, Samy Bengio, Moritz Hardt, Benjamin Recht, and Oriol Vinyals.
\newblock Understanding deep learning requires rethinking generalization.
\newblock In {\em 5th International Conference on Learning Representations,
  {ICLR} 2017, Toulon, France, April 24-26, 2017}, 2017.

\bibitem{zhao2021quantitative}
Yang Zhao and Hao Zhang.
\newblock Quantitative performance assessment of cnn units via topological
  entropy calculation.
\newblock In {\em International Conference on Learning Representations}, 2021.

\bibitem{zhao2022penalizing}
Yang Zhao, Hao Zhang, and Xiuyuan Hu.
\newblock Penalizing gradient norm for efficiently improving generalization in
  deep learning.
\newblock In {\em International Conference on Machine Learning}, pages
  26982--26992. PMLR, 2022.

\bibitem{DBLP:journals/corr/abs-1806-02891}
Bolei Zhou, Yiyou Sun, David Bau, and Antonio Torralba.
\newblock Revisiting the importance of individual units in cnns via ablation.
\newblock {\em arXiv preprint}, abs/1806.02891, 2018.

\end{thebibliography}
}
\appendix

\section{Experiments on ImageNet}

\paragraph{Dataset and networks.} In this section, the classification task is performed on the ImageNet dataset by using the same VGG16 architecture. Five networks are trained from scratch and eventually have different generalization ability. Without randomly corrupting labels, here we only alter the training strategies like momentum or dropout for changing the networks. Since classification on ImageNet is commonly supposed as a more difficult task, it is harder for networks to reach zero training error. Table \ref{table: imagenet accuracy} shows the training and testing accuracies of all the 5 networks used in our experiments, where the generalization gap ranges from 0.054 to 0.564.

\begin{table}[htt]
    \begin{center}
    \begin{tabular}{lccc}
    \toprule
    Model & Training Acc & Testing Acc & Gap\\
    \midrule
    Model A & 0.732 & 0.657 & 0.075\\
    Model B & 0.730  &  0.600 & 0.130 \\   
    Model C & 0.818 & 0.543 & 0.275 \\   
    Model D & 0.828 & 0.444 & 0.384\\    
    Model E & 0.978 & 0.374 & 0.604 \\   
    \bottomrule
    \end{tabular}
    \end{center}
    \caption{Training and testing accuracies of 5 networks.}
    \label{table: imagenet accuracy}
    \vspace{-0.18in}
\end{table}

\begin{figure*}[t]
    \centering
    \includegraphics[width = 1\columnwidth]{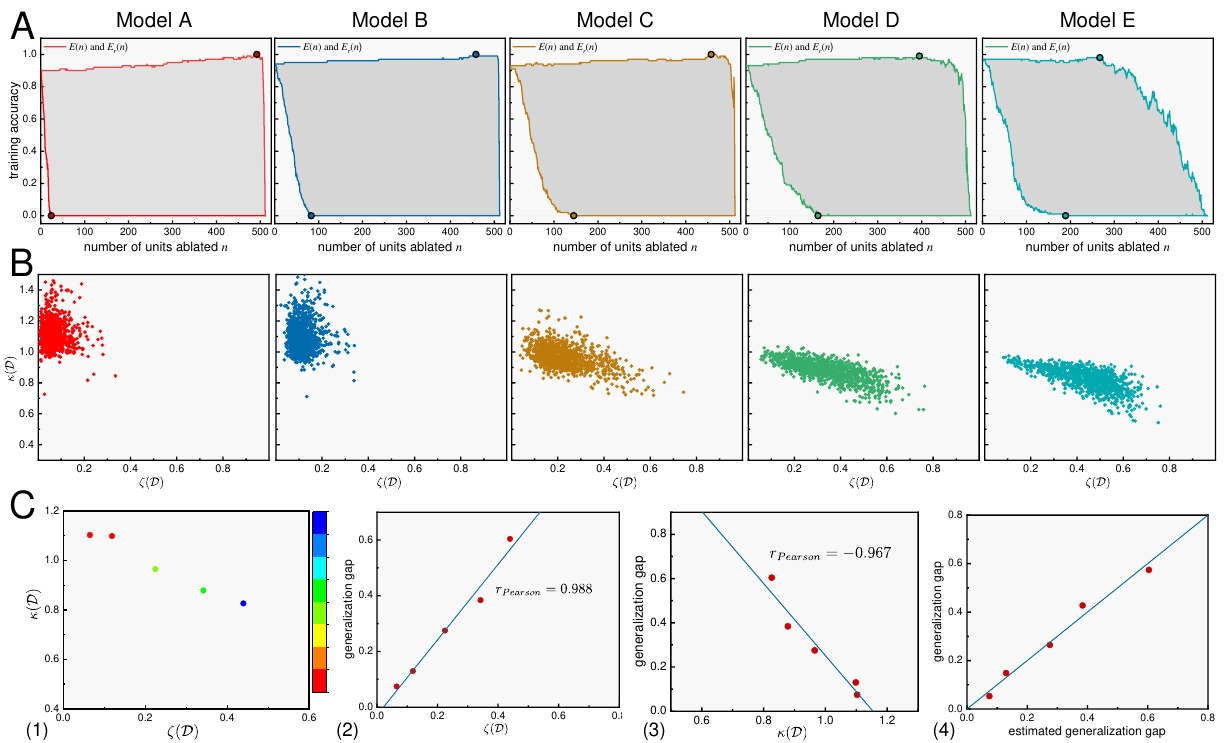}
    \caption{(A) The evolution example curves of accuracy $E(n, \mathcal{D}_j)$ and $E_r(n, \mathcal{D}_j)$ for the five networks. (B) Scatter plot between the two quantities $\zeta(\mathcal{D})$ and $\kappa(\mathcal{D})$ across all the classes in the ImageNet dataset. (C) Scatter plots between different quantities.}
    \label{fig: vgg results}
\end{figure*}

\paragraph{Results.} Similarly, the cumulative unit ablation is performed for each network on all the class in the dataset at first. Fig.\ref{fig: vgg results}A shows almost the same tendency of the turning points ($n_0(\mathcal{D}_j)$ and $n_0^{r}(\mathcal{D}_j)$) and accuracy curve ($E(n, \mathcal{D}_j)$ and $E_r(n, \mathcal{D}_j)$) as the result on CIFAR100.

Next, $\zeta(\mathcal{D})$ and $\kappa(\mathcal{D})$ are calculated based on the two curves across all the classes. Fig.\ref{fig: vgg results}B presents the scatter plot of all the quantity pairs $(\zeta(\mathcal{D}_j), \kappa(\mathcal{D}_j))$ . We could see that as the generalization ability of networks becomes worse, their quantity pairs move gradually towards the bottom right direction. Besides, we could also find that the quantity pairs of those networks with better generalization would be more gathered, especially with respect to the $\zeta(\mathcal{D})$.

Then, Fig.\ref{fig: vgg results}C(1) shows the similar scatter plot of the quantity pair $(\zeta(\mathcal{D}), \kappa(\mathcal{D}))$ for the five networks. The colors are in the same scale with it used in Fig.\ref{fig: curve fitting}A. As we could see that from the top right corner to the left bottom, the generalization gap gradually increases. This is the same with the regularity in in Fig.\ref{fig: curve fitting}A, and again provides evidence in the correlation between the two sparsity quantities and network generalization gap.

Fig.\ref{fig: vgg results}C(2) visualizes the linear relation between $\zeta(\mathcal{D})$ and generalization gap for five networks. $\zeta(\mathcal{D})$ has an extremely strong linear correlation with generalization gap, where the Pearson correlation coefficient reaches remarkably $0.998$. Even use $\kappa(\mathcal{D})$ instead of $\zeta(\mathcal{D})$, the results in Fig.\ref{fig: vgg results}C(3) still shows a well degree of linear correlation with Pearson correlation $0.967$. This support strongly that $\zeta(\mathcal{D})$ and $\kappa(\mathcal{D})$ are both really effective characterizations of the generalization of DNNs.

Fig.\ref{fig: vgg results}C(4) shows that the points with respect to five networks lies very closely to the reference line, indicating that the estimated generalization gap and the true gap are almost equal. Also, the SSR here is only 0.004.

\section{Experiments with MobileNet on CIFAR100}

For this experiment, we use the MobileNet architecture (Howard AG et al. 2017) for classifying CIFAR100. When training the networks, we still partially corrupt the dataset with the same fractions as being used in the VGG16. Similarly, to get networks with different generalization gaps, we use the same training strategies as them in VGG16 except for the dropout and batch normalization. 

The results are presented in Fig.\ref{fig: mobilenet}. As expected, we could see in Fig.\ref{fig: mobilenet}A that when the fraction of corrupted labels becomes higher, $n_0(\mathcal{D}_j)$ becomes gradually larger while $n_0^{r}(\mathcal{D}_j)$ becomes smaller, and meanwhile, the area becomes smaller as well. This is the same with VGG16 (Fig.3A in the paper). Then, Fig.\ref{fig: mobilenet}B shows the quantity pair $(\zeta(\mathcal{D}), \kappa(\mathcal{D}))$ of each network with a scatter plot. The quantity pair of networks with better generalization ability mostly lie in the top right corner, and contrarily these with poor generalization ability lie in the bottom left corner. By using our estimation model, Fig.\ref{fig: mobilenet}C shows the estimated generalization gap of all the networks. We could see that the points are scattered closely between the reference line $x = y$. In the meantime, the SSR of this fitting is 0.074, which demonstrates the effectiveness of our model.

\section{Experiments with ResNet34 on CIFAR100}

For this experiment, we change the network architecture to ResNet34 (He et al. 2016) for classifying CIFAR100. When training the networks, all the strategies are the same with them used in MobileNet for obtaining the networks with diverse generalization gaps.

\begin{figure*}[htbp]
    \centering
    \includegraphics[width = 1\columnwidth]{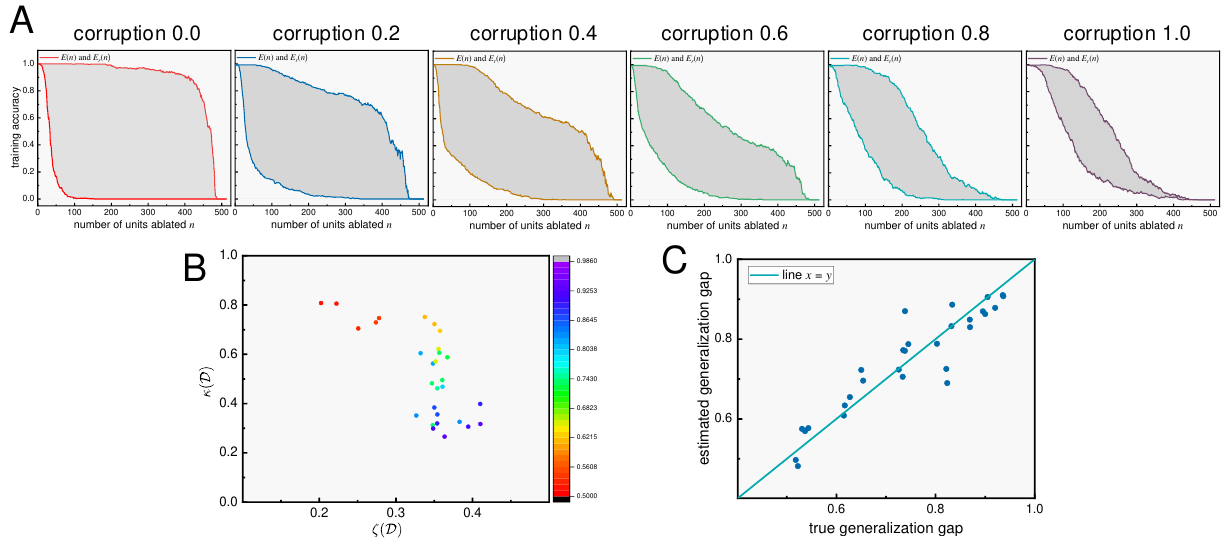}    
    \caption{Results on MobileNet. (A) The evolution example curves of accuracy $E(n, \mathcal{D}_j)$ and $E_r(n, \mathcal{D}_j)$ on dataset with separate corruption fraction of labels in $\{0, 0.2, 0.4, 0.6, 0.8, 1.0\}$. (B) Scatter plot between the two key quantities $\zeta(\mathcal{D})$ and $\kappa(\mathcal{D})$ across all the networks. The color of each point indicates the generalization gap, where red represents the smallest value and purple represents the largest. (C) Scatter plot between the estimated generalization gap and the true generalization gap.}
    \label{fig: mobilenet}
\end{figure*}

\begin{figure*}[htbp]
    \centering
    \includegraphics[width = 1\columnwidth]{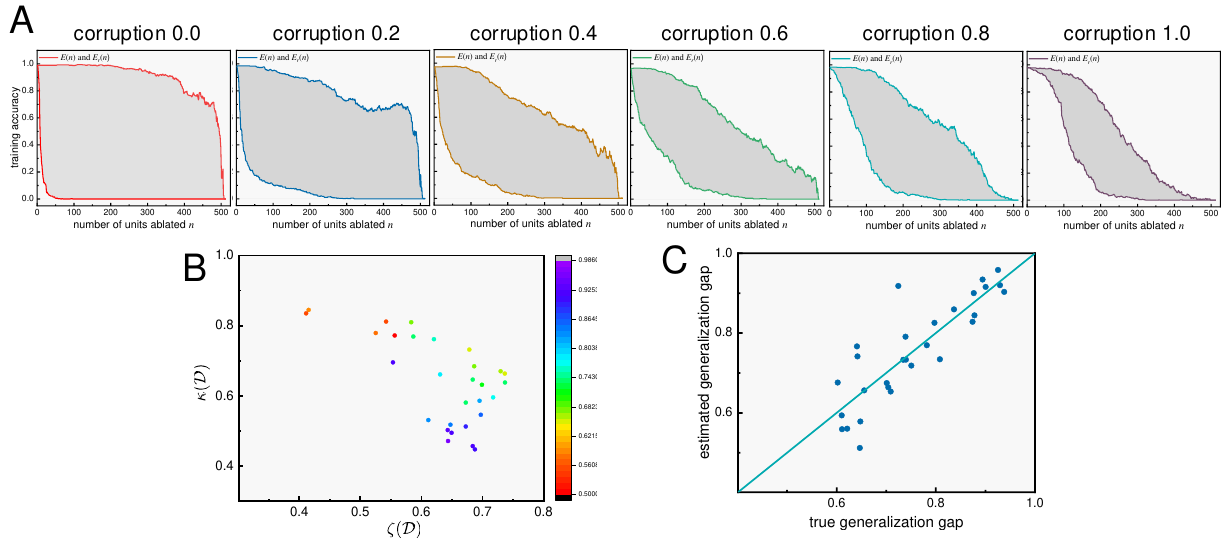}
    \caption{Results on ResNet34. The meaning of all the figures is the same with Fig.\ref{fig: mobilenet}.}
    \label{fig: resnet}
\end{figure*}

Fig.\ref{fig: resnet} presents the results. All the figures in the Fig.\ref{fig: resnet} are in the same meaning with Fig.\ref{fig: mobilenet}. And we get the similar results with the results when using VGG16 and MobileNet. The SSR here is 0.122, which shows again the effectiveness of our model.

\vspace{1in}

\section{Implementation Details of Experiments}

\paragraph{Implementation with VGG16 on CIFAR100.} In this experiment, the networks we used are the standard VGG16 architecture. To get networks with a wider range of generalization, we perform the following implementations when training,

\begin{itemize}
    \item Build network with or without batch normalization.
    \item Use dropout at the fully connected layers with rate from $\{0, 0.3, 0.5\}$.
    \item Use SGD optimizer with momentum from $\{0, 0.5, 0.9\}$.
    \item Use L2 regularization with coefficient from $\{0, 0.0001\}$.
    \item Use batch size from $\{128, 256, 512\}$.
    \item Use or not use data augmentation with random cropping, horizontal flipping and rotation.
    \item Partially corrupt labels in the training dataset with fractions from $\{0, 0.2, 0.4, 0.6, 0.8, 1.0\}$.
\end{itemize}

\paragraph{Implementation with VGG16 on ImageNet.} In this experiment, the networks we used are still the standard VGG16 architecture. And the following implementations are performed when training,

\begin{itemize}
    \item Model A. The hyper-parameters are the same as those in the paper (Simonyan and Zisserman 2015).
    \item Model B. The hyper-parameters are the same as those in Model A, except for not using the data augmentation strategy.
    \item Model C. The hyper-parameters are the same as those in Model B, except for changing the momentum to 0.
    \item Model D. The hyper-parameters are the same as those in Model C, except for that only the first fully connected layer use the dropout with the rate of 0.3.
    \item Model E. None of the conventional training enhancement technique is applied. Basically, It is Model D without using dropout and l2 regularization.

\end{itemize}

\paragraph{Units in cumulative unit ablation.}
\noindent For our cumulative unit ablation, we choose units at the "block5{\_}conv3" layer, which is the last convolution layer in VGG16 architecture.

\section{Code Release}

The code is available at github.

https://github.com/zhaoyang-0204/generalization-estimating.git

\end{document}